\documentclass[conference]{IEEEtran}
\IEEEoverridecommandlockouts

\usepackage{cite}
\usepackage{amsmath,amssymb,amsfonts}
\usepackage{algorithmic}

\usepackage{graphicx}
\usepackage{subcaption}
\usepackage{bbding}
\usepackage{epsfig}
\usepackage{booktabs} 
\usepackage{cite}
\usepackage{hyperref}
\usepackage{textcomp}
\usepackage{xcolor}
\def\BibTeX{{\rm B\kern-.05em{\sc i\kern-.025em b}\kern-.08em
    T\kern-.1667em\lower.7ex\hbox{E}\kern-.125emX}}
    
\def\etal{\emph{et al}. }

\begin{document}

\title{The Demon is in Ambiguity: Revisiting  Situation Recognition with Single Positive Multi-Label Learning\\
}

\author{\IEEEauthorblockN{1\textsuperscript{st} Yiming Lin}
\IEEEauthorblockA{\textit{School of Advanced Technology} \\
\textit{Xi'an Jiaotong-Liverpool University}\\
Suzhou, China \\
yiming.lin21@student.xjtlu.edu.cn}
\and
\IEEEauthorblockN{2\textsuperscript{nd} Yuchen Niu}
\IEEEauthorblockA{\textit{School of Advanced Technology} \\
\textit{Xi'an Jiaotong-Liverpool University}\\
Suzhou, China \\
Yuchen.Niu17@alumni.xjtlu.edu.cn}
\and
\IEEEauthorblockN{3\textsuperscript{rd} Shang Wang}
\IEEEauthorblockA{\textit{School of Advanced Technology} \\
\textit{Xi'an Jiaotong-Liverpool University}\\
Suzhou, China \\
shang.wang24@student.xjtlu.edu.cn}
\and
\IEEEauthorblockN{4\textsuperscript{th} Kaizhu Huang}
\IEEEauthorblockA{\textit{Data Science Research Center} \\
\textit{Duke Kunshan University}\\
Suzhou, China \\
kaizhu.huang@dukekunshan.edu.cn}
\and
\IEEEauthorblockN{5\textsuperscript{th} Qiufeng Wang}
\IEEEauthorblockA{\textit{School of Advanced Technology} \\
\textit{Xi'an Jiaotong-Liverpool University}\\
Suzhou, China \\
qiufeng.wang@xjtlu.edu.cn}
\and
\IEEEauthorblockN{6\textsuperscript{th} Xiao-Bo Jin}
\IEEEauthorblockA{\textit{School of Advanced Technology} \\
\textit{Xi'an Jiaotong-Liverpool University}\\
Suzhou, China \\
xiaobo.jin@xjtlu.edu.cn}
}

\maketitle

\begin{abstract}

Context recognition (SR) is a fundamental task in computer vision that aims to extract structured semantic summaries from images by identifying key events and their associated entities. Specifically, given an input image, the model must first classify the main visual events (verb classification), then identify the participating entities and their semantic roles (semantic role labeling), and finally localize these entities in the image (semantic role localization). Existing methods treat verb classification as a single-label problem, but we show through a comprehensive analysis that this formulation fails to address the inherent ambiguity in visual event recognition, as multiple verb categories may reasonably describe the same image. This paper makes three key contributions: First, we reveal through empirical analysis that verb classification is inherently a multi-label problem due to the ubiquitous semantic overlap between verb categories. Second, given the impracticality of fully annotating large-scale datasets with multiple labels, we propose to reformulate verb classification as a single positive multi-label learning (SPMLL) problem - a novel perspective in SR research. Third, we design a comprehensive multi-label evaluation benchmark for SR that is carefully designed to fairly evaluate model performance in a multi-label setting. 
To address the challenges of SPMLL, we futher develop the Graph Enhanced Verb Multilayer Perceptron (GE-VerbMLP), which combines graph neural networks to capture label correlations and adversarial training to optimize decision boundaries. Extensive experiments on real-world datasets show that our approach achieves more than 3\% improvement on the more meaningful multi-label Average Precision (MAP) metric while remaining competitive on traditional top-1 and top-5 accuracy metrics. To our knowledge, our research is the first work that the formulate, solving, and evaluating of verb classification in the SPMLL fashion, which provides theoretical insights and practical tools for advancing situation recognition research. 
\end{abstract}


\section{Introduction}
\label{sec:intro}
\begin{figure}[t] 
  \centering
   \includegraphics[width=1.0\linewidth]{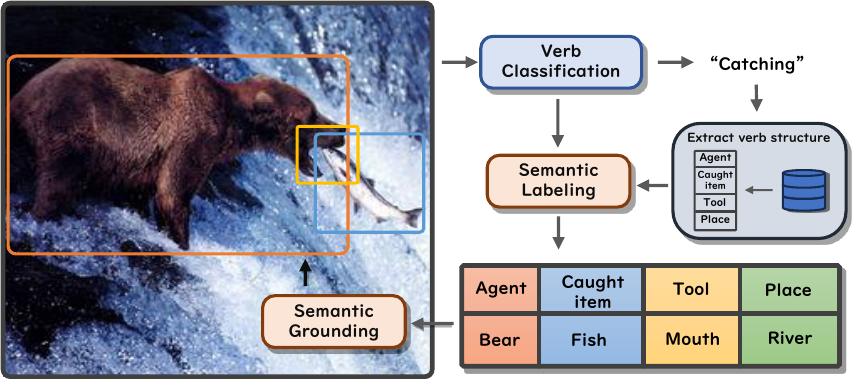}
   \caption{Illustration of Situation Recognition. 
    Given an image, the model first classifies the type of “Catching” event in the image, then labels noun phrases ``Bear", ``Fish", ``Mouth", and “River” according to the semantic roles of “Catching” event (``Agent", ``Caught Item", ``Tool", ``Place"), and finally regresses the bounding box of each classified visual object.}
    \label{fig:SR_sample_v2}
\end{figure}

Modern multimedia applications increasingly demand systems that can understand images at both the object level (recognizing individual entities) and the event level (comprehending interactions and activities). Situation Recognition (SR) has emerged as a crucial task addressing this need by extracting structured semantic representations from images \cite{yatskar2016situation, pratt2020grounded}. 
Generally, SR \cite{yatskar2016situation, pratt2020grounded} can be decomposed into three interrelated sub-tasks: verb classification, semantic role labeling, and semantic role grounding. Fig.~\ref{fig:SR_sample_v2} shows a typical example of SR. Given an image, verb classification requires classifying the occurred visual event types (known as ``verbs"). Then, the semantic role labeling intends to classify the noun phrases according to the event-specific arguments (also known as ``roles"). Further, the semantic role grounding task focuses on regressing the corresponding bounding boxes for each visual object.


Among the above three sub-tasks, verb classification is relatively independent yet serves as the bottleneck in SR since it determines the specific roles for the following labeling and grounding tasks. Recent advances have primarily focused on improving verb classification, underscoring its pivotal role in SR ~\cite{roy2024clipsitu, li2022clip}. However, current approaches treat such problems in a multi-class classification fashion, which we argue fundamentally misrepresents the inner nature of visual event recognition.

\begin{figure}[t]
    \centering
    \includegraphics[width=1.0\linewidth]{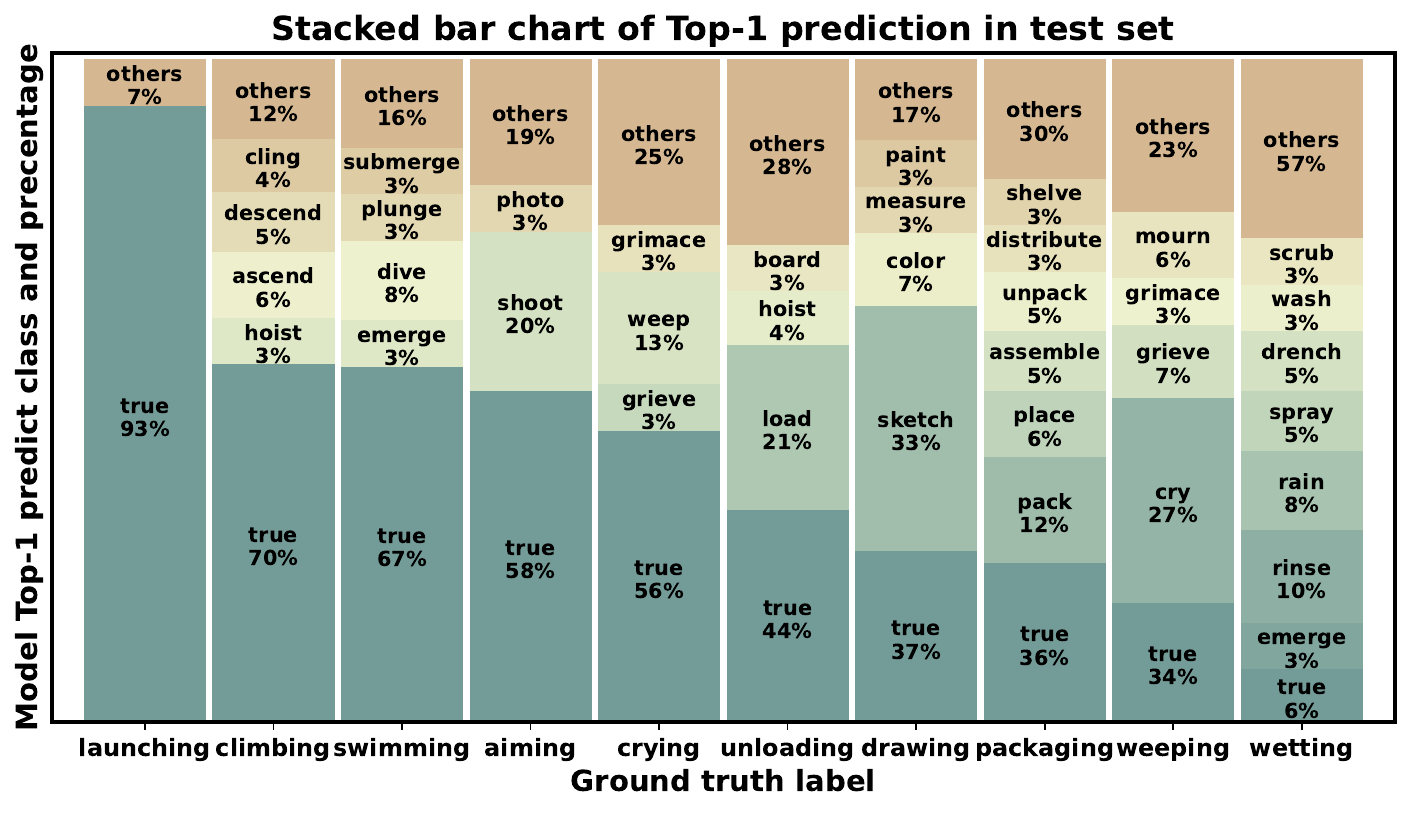}
    \caption{
    Illustration of Baseline model's Top-1 predictions on test set for random sampled 10 classes. Each bar shows the predicted label and its proportion, with the ground truth label on x-axis. It can be seen that most wrong cases are semantically similar, or even semantically overlapped with the ground truth class.
    }
    \label{fig:enter-label}
\end{figure}

\begin{figure*}[h]
  \centering
  \begin{subfigure}{0.49\linewidth}
    \includegraphics[width=1.0\textwidth]{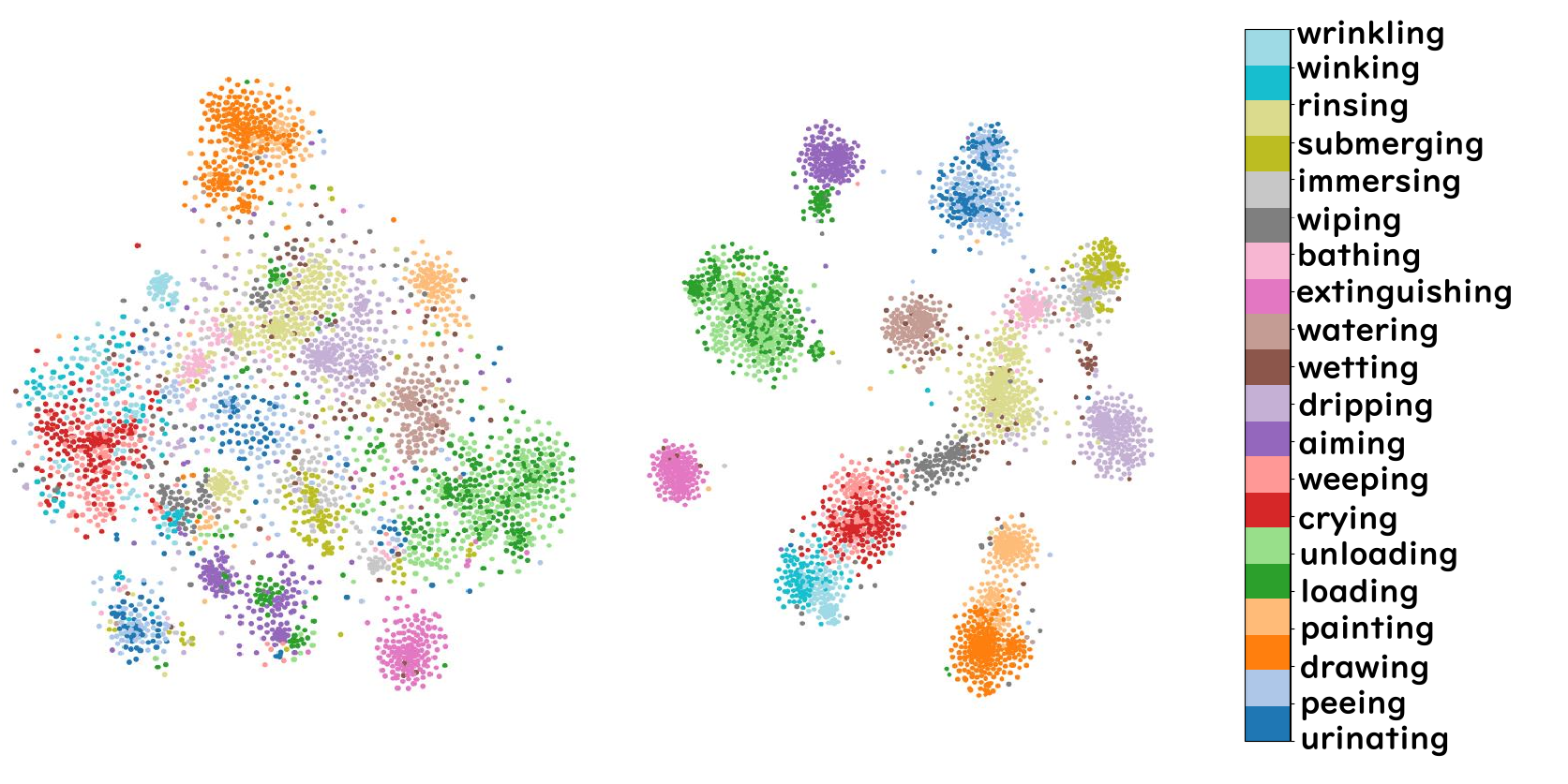}
    \caption{Exploring the ambiguity problem from the perspective of model embeddings. Comparing visualizations of CLIP embeddings (left) and embeddings before the classification head (right), we find that the essence of the ambiguity problem is the severely overlapping class distributions}
    \label{fig:toy_tsne_with_original_v2}
  \end{subfigure}
  \hfill
  \begin{subfigure}{0.49\linewidth}
    \includegraphics[width=1.0\textwidth]{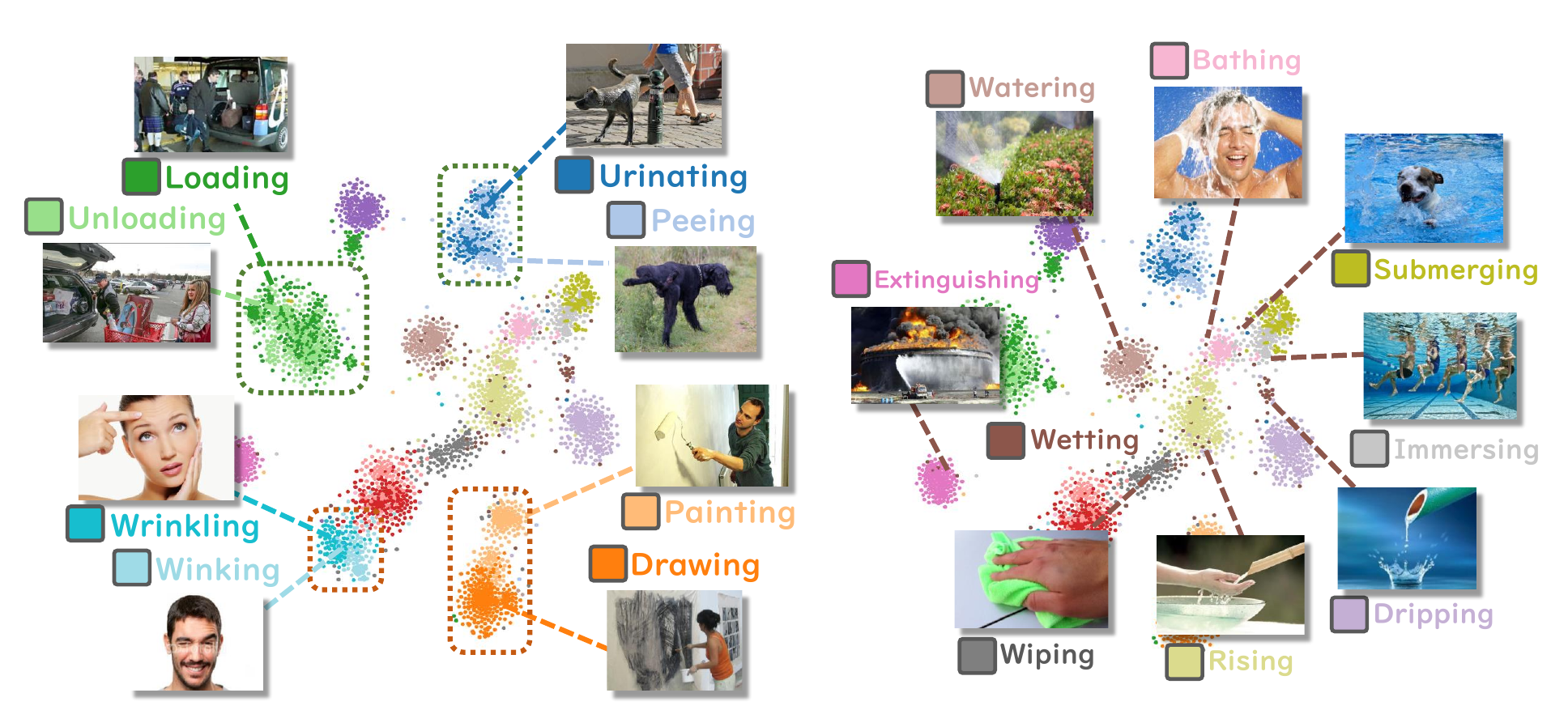}
    \caption{Two different types of overlapping distributions (left) and examples of blurred images (right). The green and red boxes in the left image show the cases of full overlap and partial overlap, respectively. The right image shows a typical example of multiple classes overlapping; the ``wetting" class is heavily overlapped with multiple related classes.
    }
    \label{fig:two_case_and_wet_case}
  \end{subfigure}
  \caption{Illustration of sample embeddings with T-SNE visualization. Different colors represent samples of different categories.}
  \label{fig:short}
\end{figure*}

Through extensive empirical analysis, we identify ambiguity as a pervasive yet understudied challenge in verb classification. 
From Fig.~\ref{fig:enter-label}, we first notice the model's wrong cases in the Top-1 prediction are semantically overlapped with the ground truth classes.
To gain more insight, we plot out the latent space for different class samples in Fig.~\ref{fig:toy_tsne_with_original_v2} by employing t-SNE \cite{van2008visualizing}. 
It can be observed that both the CLIP \cite{radford2021learning} embeddings and the embeddings before the classification head exhibit substantial overlap among many different classes. A typical example is the overlap between “crying” and “weeping,” highlighted in red and pink in Fig.~\ref{fig:toy_tsne_with_original_v2}.

Taking a closer look at this issue, in Fig.~\ref{fig:two_case_and_wet_case}, we illustrate that the class overlap can be quite complex. On the left of Fig.~\ref{fig:two_case_and_wet_case}, we highlight class pairs (e.g., ``drawing" \textit{vs.} ``painting" in the red box) that may have similar but subtly distinct meanings, making it difficult to establish a clear boundary between them. Furthermore, in extreme cases (e.g., ``loading" \textit{vs.} ``unloading"), the overlapping classes are deeply intertwined, suggesting that an image could reasonably belong to multiple labels. On the right of Fig.~\ref{fig:two_case_and_wet_case}, we present a more intricate scenario where multiple classes partially overlap or are encompassed within a specific class. For instance, the ``wetting" class on the right overlaps with several other classes due to shared characteristics. 

These observations reveal that the classes in the verb classification task are not mutually exclusive (e.g., a classroom scene might simultaneously involve ``teaching", ``lecturing", ``studying", and ``writing"), rendering the traditional multi-class classification training and evaluation schemes inadequate. 




Inspired by the above analysis, we propose that verb classification should not be narrowly treated as a multi-class problem but instead approached as a multi-label classification task.
However, training and evaluating from a multi-label perspective is hindered by the reality that only a single label is explicitly assigned to each sample in existing verb classification datasets.
While annotating the entire large-scale imSitu training set~\cite{yatskar2016situation}\footnote{To our knowledge, imSitu~\cite{yatskar2016situation}\cite{pratt2020grounded} is the only publicly available benchmark dataset in SR annotated using a multi-class classification scheme, making it cost-prohibitive to annotate in a multi-label manner.} with 504 classes in a multi-label fashion would be prohibitively expensive and impractical, we suggest that the verb classification problem can be viewed and addressed as an instance of the recently proposed Single Positive Multi-Label Learning (SPMLL)~\cite{pratt2020grounded} framework, which discusses training a multi-label classification model using only one positive label per instance. 

To evaluate multi-label models trained using the SPMLL scheme, we first curate a mini-benchmark dataset by carefully selecting ten categories from the original imSitu benchmark~\cite{yatskar2016situation} and re-annotating these samples in a multi-label manner. The purpose of the mini-benchmark is to help us validate the existence of ambiguous problems and further enable us to evaluate any pre-annotation pipelines that help to scale the evaluation benchmark with less effort. Finally, with the help of our proposed VLM-LLM pre-annotation pipeline, we develop a novel multi-label evaluation benchmark with 25,200 annotated images that can assess both traditional single-label metrics (top-1/5 accuracy) and multi-label performance (MAP). To our knowledge, this is the first large-scale multi-label evaluation dataset in the SR literature.





On the algorithmic side, adopting a multi-label classification framework has allowed us to leverage label correlations to enhance model performance. To this end, we develop a novel method called Graph-Enhanced Verb MLP (GE-VerbMLP), which integrates a graph convolutional network (GCN)~\cite{kipf2016semi} to incorporate label correlations into the classifier’s architecture seamlessly. Instead of using complex manual correlation matrix designs, we utilize semantic similarity as a proxy for ``pseudo" label correlation.
For scenarios where class distinctions are subtle with obscure decision boundaries, we employ adversarial training~\cite{qian2022survey} to generate more challenging examples, forcing the model to learn smoother decision boundaries and improve its generalization capabilities.

Our experimental results demonstrate 3+\% improvements in MAP while maintaining competitive single-label accuracy, validating that addressing ambiguity leads to more robust situation understanding.

In a nutshell, our work makes four main contributions to the field of situation recognition:
\begin{itemize}
    \item  We provide the first comprehensive analysis of ambiguity in verb classification through embedding visualization and human annotation studies.
    \item We pioneer the reformulation of verb classification as an SPMLL problem, better matching the true nature of visual event recognition.
    \item We combine graph neural networks and adversarial learning to achieve state-of-the-art performance.
    \item We generate a large-scale multi-label dataset, which is the first comprehensive benchmark to enable proper assessment of SR methods.
\end{itemize}

\section{Relate Work}
\label{sec:literature}


\subsection{Situation Recognition}
Yatskar \etal \cite{yatskar2016situation} first introduced a model based on conditional random fields and a tensor combination method. As an extension of situation recognition, Pratt \etal. \cite{pratt2020grounded} proposed the concept of Grounded Situation Recognition (GSR) and two basic models, Independent Context Localizer (ISL) and Joint Context Localizer (JSL), and developed the SWiG dataset. Following exploration~\cite{mallya2017recurrent, li2017situation, suhail2019mixture, cooray2020attention} demonstrated, the SR problem can be optimized with recurrent neural networks, graph neural networks, and top-down attention.
Further, Cho \etal. \cite{cho2021grounded} leverage a Transformer-based encoder-decoder structure for semantic labeling and grounding tasks, and CoFormer~\cite{cho2022collaborative} improved the accuracy of verb prediction by introducing the glance and gaze transformers. Recent works~\cite{roy2024clipsitu, li2022clip} focus on large-scale vision-language pre-trained models~\cite{radford2021learning}, showing it is effective on SR tasks. Specifically, CLIP-Event~\cite{li2022clip} designed a specific pre-training mechanism on the CLIP model for the SR task, and the ClipSitu model~\cite{roy2024clipsitu} verified that training directly with a multi-layer perceptron on the parameter-frozen CLIP model achieves the best results. Recently, a pioneering study~\cite{sanders2022ambiguous} has addressed the ambiguous problem in verb classification tasks, and they propose the SQUID-E datasets designed to evaluate the uncertainty of human annotations for each image. However, such exploration is still based on the assumption that an image only belongs to a single class annotation, while the view of annotating multiple labels per image is still unexplored. 

\subsection{Single Positive Multi-label Learning}

As a extreme case of weakly supervised multi-label learning (WSML), Single Positive Multi-Label Learning (SPMLL) addresses scenarios where each training sample has only one annotated positive label, with the rest either missing or potentially being false negatives. This challenging setting was first formally introduced by \cite{cole2021multi}, who proposed several foundational techniques to mitigate the impact of unobserved labels. Their early solutions included the use of weak negative samples to approximate missing negatives, label smoothing to reduce overconfidence in uncertain labels, and online label estimation to dynamically refine predictions during training.

In the following works \cite{kim2022large, kim2023bridging, xu2022one, zhou2022acknowledging}, modifying the loss function has proven to be an effective measure of negative pseudo-labels and can be further optimized by leveraging model memory \cite{kim2022large} and class activation mapping (CAM)~\cite{kim2023bridging}. In addition, the self-paced loss correction (SPLC) method \cite{zhang2021simple} utilized confidence scores to generate pseudo-labels. Another work names Fundus SPMLL (FSP)~\cite{hu2025co} perform co-pseudo labeling with active selection in fundus images by dynamically adjusting pseudo-label thresholds and selecting high-confidence samples. Beyond utilizing pseudo-labeling methods to alleviate the SPMLL problem, the semantic correspondence prompt network (SCPNet)~\cite{ding2023exploring} presents a novel approach utilizing semantic associations to improve model performance, its concurrent work HSPNet \cite{wang2023hierarchical} designed a hierarchical conditional prompt (HCP) to explore the inherent label-group dependency and proposed a hierarchical graph convolutional network (HGCN) to refine label features and semantic representations. Meanwhile, Zhang~\etal developed a semantic-guided representation learning approach (SigRL)~\cite{zhang2025semantic} that captures multi-label correlations via graph structures and enhances visual-textual alignment for zero-shot recognition. For data augmentation perspective, SpliceMix \cite{wang2025splicemix} proposed a semantic-preserving blending strategy for multi-label images.



\subsection{Adversarial Training}
Goodfellow \etal\cite{goodfellow2014explaining} proposed adversarial training to improve model robustness, providing a foundational approach Fast Gradient Sign Method (FGSM). In subsequent studies, researchers commonly utilize gradient-based methods, such as Projected Gradient Descent (PGD) \cite{madry2017towards}. Additionally, novel methods implementing generative adversarial networks (GAN)~\cite{goodfellow2014generative} and diffusion models (DDPM) \cite{ho2020denoising} have been proposed to generate adversarial samples. In addition to improving the model's robustness, recent studies have demonstrated that adversarial training can effectively achieve boundary smoothing~\cite{qian2022survey}. This paper reveals the effectiveness of classical FGSM and PGD methods in a multi-label verb classification scheme.
\section{Problem Formulation}
\label{sec:problem formulation}
In this section, we formally introduce how to frame the verb classification problem as an SPMLL problem.
Assume that $\mathcal{X}$ represents a $d$-dimensional sample space and $\mathcal{Y} = \{1,2,\cdots,L\}$ represents a label space containing $L$ class labels. 
The multi-label classification task is to learn a function $h: \mathcal{X} \to [0,1]^{L}$ from a multi-label dataset $\mathcal{D} = \{(x_i,y_i)\}_{i = 1}^m$, where $x_i \in \mathcal{X}$ is a $d$-dimensional feature vector and $y_i \in \{0,1\}^L$ ($y_{ik} = 1$ if $x_i$ belongs to class $k$) is the label vector attached to $x_i$. The objective of multi-label classification is to find a parameterized function $f_{\theta}$ that minimizes the following empirical risk loss
\begin{equation}
    R_{\textrm{full}}(f_{\theta}) = \frac{1}{m}\sum_{i = 1}^m L(f_{\theta}(x_i),y_i),
\end{equation}
where $L$ can be multiple binary cross entropy loss functions (or multiple binary focal loss functions), and $f$ is any function mapping from $R^d$ to $\{0,1\}^L$.

Ideally, when we minimize the empirical risk, we should be able to fully observe the label vector or ground truth of the sample. However, in the current literature of the verb classification task, each training sample is only annotated with one positive label, although it is essentially a multi-label problem. In this extreme case, the verb classification can be formulated as an SPMLL problem, where the labels of all training samples are degenerated into a one-hot encoding $z_i$
\begin{equation}
z_{ik} = \begin{cases}
1, & k \textrm{ is one of all positive labels of } x_i, \\
0, & \textrm{otherwise}.
\end{cases}
\end{equation}
Based on the above setting, we redefine the empirical risk loss for SPMLL problem:
\begin{equation}
R_{\textrm{partial}}(f_{\theta}) = \frac{1}{m}\sum_{i = 1}^m L(f_{\theta}(x_i),z_i).
\end{equation}
Intuitively, as we get more information about the labels, we should be able to get a smaller empirical risk loss, i.e.,$R_{\textrm{full}}(f_{\theta}) \le R_{\textrm{partial}}(f_{\theta})$.

\section{Multi-label Benchmark Collection}
\label{sec:dataset}
\begin{figure}[htbp]
\centering
\includegraphics[width=0.45\textwidth]{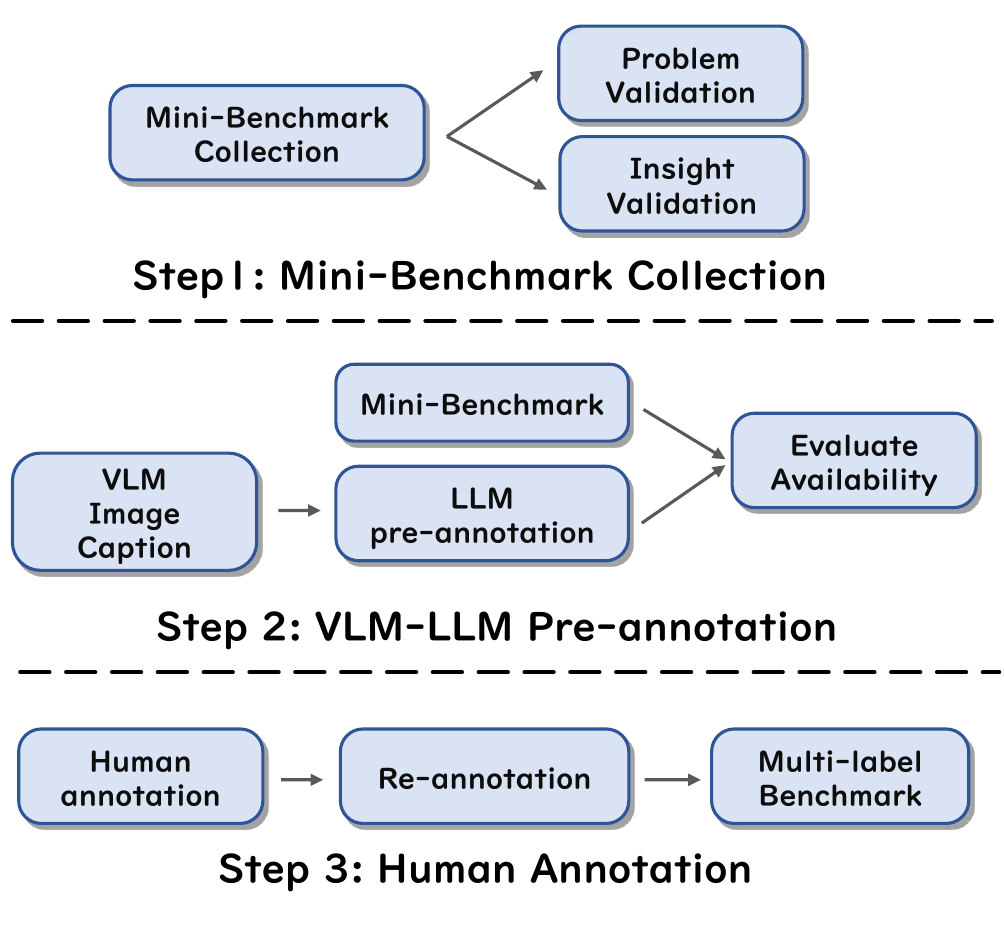}
\caption{
Three main steps for data collection: 1) We gather a mini-benchmark dataset to validate the problem's existence and derive key algorithmic insights; 2) we employ large language models (LLMs) for pre-annotation by extracting image captions and using them as LLM's input to generate initial labels; 3) we refine the annotations through human verification and expert re-annotation, resulting in a high-quality multi-label benchmark dataset.}
\label{fig:dataset_step}
\end{figure}

In order to validate the ambiguity problem and enable the evaluation of the model's performance under the SPMLL setting, we distill and re-annotate a multi-label benchmark from the imSitu dataset by three steps in Fig. \ref{fig:dataset_step}.

\subsection{Mini-Benchmark Collection}
The verb category comprises 504 classes, making complete manual annotation for all imSitu data impractical. Accordingly, we first manually annotated a mini-benchmark with less effort to validate the existence of the ambiguity issue. Subsequently, such a mini-benchmark is used to verify the feasibility of following the Vision Language Model-Large Language Model (VLM-LLM) pre-annotation step.

\subsubsection{Class Selection and Labeling Process}
We select and re-annotate ten classes from the original imSitu dataset to create the mini-benchmark. To ensure fairness in class subset selection, we first rank all classes in the original dataset based on their accuracy in the ClipSitu Verb MLP model~\cite{roy2024clipsitu} and group them into ten bins. From each bin, we randomly select one class for re-annotation. The selected classes and their top-1 accuracy performance are depicted in the left part of Tab.~\ref{tab:label_static_and_SR_benchmark}. During the annotation process, we hire two expert annotators to refine the multi-label annotations from a total of 504 classes.
To ensure annotation accuracy, the final results were validated by another expert in the field. Finally, 500 images (50 samples per class) are annotated in a multi-label fashion.

\subsubsection{Dataset Analysis}
\begin{figure}[htbp]
    \centering
    \includegraphics[width=0.47\textwidth]{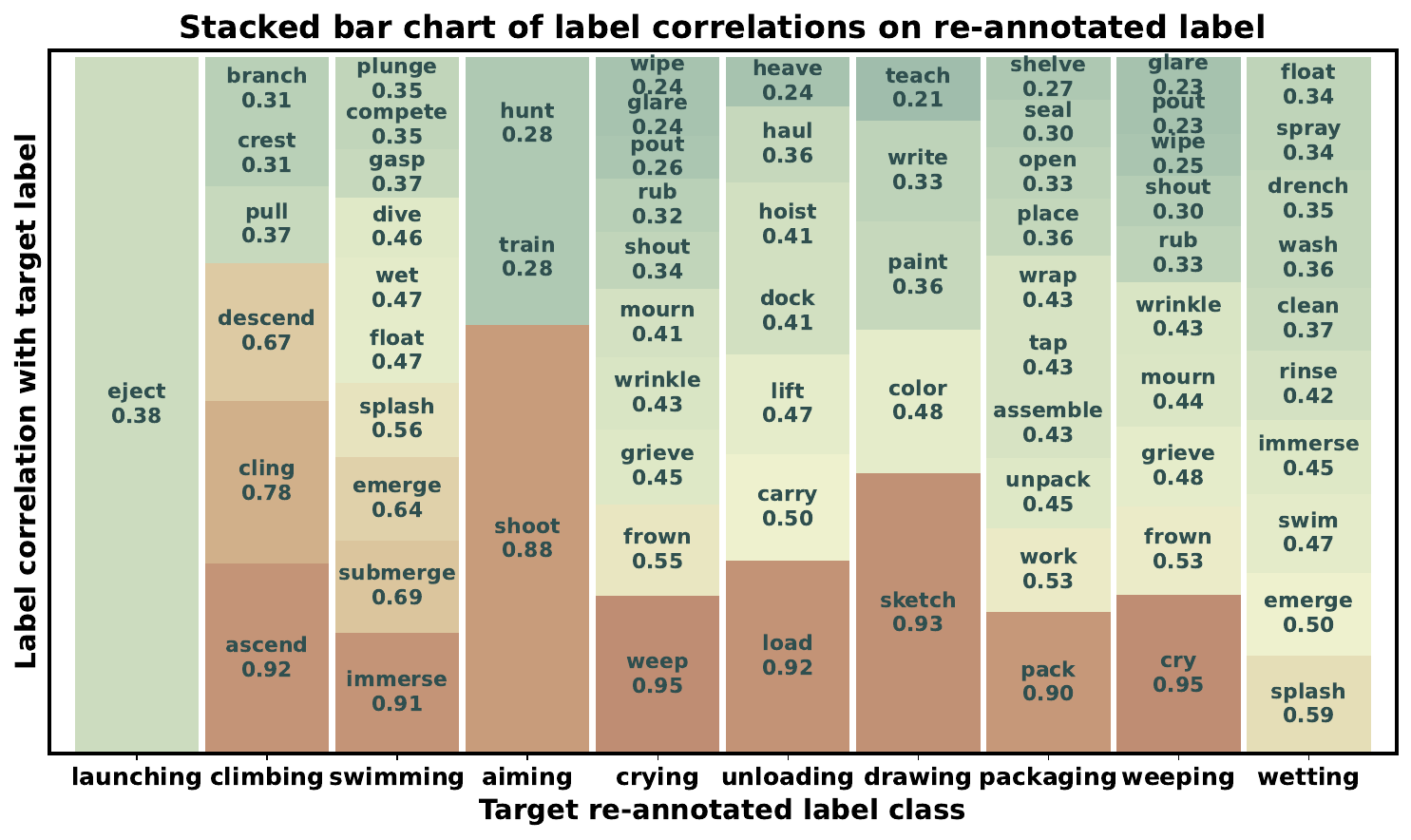}
        \caption{According to human annotation, we calculate the correlations between the ten selected categories and other labels. Each correlated label is represented by a segment in the stacked bar chart. The higher the value, the closer the color is to dark brown, which indicates the label is more likely to be annotated together with the x-axis-specified label. For illustration, we filtered out labels with correlations exceeding the threshold of $0.3$.}
    \label{fig:10_class_sep_heatmap}
\end{figure}

To illustrate the nature of the annotated multi-label dataset and provide insights into the overlapping problem, we present the label correlation of selected classes according to human annotations in Fig.~\ref{fig:10_class_sep_heatmap}.

We first validate the existence of our proposed ambiguity problem. Comparing the ground truth label correlation in Fig.~\ref{fig:10_class_sep_heatmap} with the model's top-1 predictions in Fig.~\ref{fig:enter-label}, it can be observed that human annotators recall most incorrect classes in the multi-label task setting. Moreover, compared to  Tab.~\ref{tab:label_static_and_SR_benchmark}, we find a high correlation between the accuracy and severity of label correlation. e.g. the ``launching" class has fewer correlated labels, while the ``wetting" class exhibits more severe correlations with other classes.

Further, we identify some interesting characteristics of the correlated labels: 1) Most classes are severely correlated with a few specific classes. Except for the highly ambiguous ``wetting" class, most classes are strongly related to no more than five other classes. 2) The correlated labels are semantically related to the original class. For example, ``climbing" is correlated with ``ascending", ``clinging", and ``descending". These characteristics inspire us to leverage semantic relations as a proxy for label correlation and use a k-nearest-neighbor strategy to construct a graph.

\subsection{Pre-annotation with VLM and LLM}
Due to the prohibitively high cost of manually annotating the entire test dataset (which contains 504 candidate labels per image), we propose a VLM-LLM pre-annotation pipeline to significantly reduce the number of candidate categories requiring human verification.

\begin{figure}[htbp]
\centering
\includegraphics[width=0.49\textwidth]{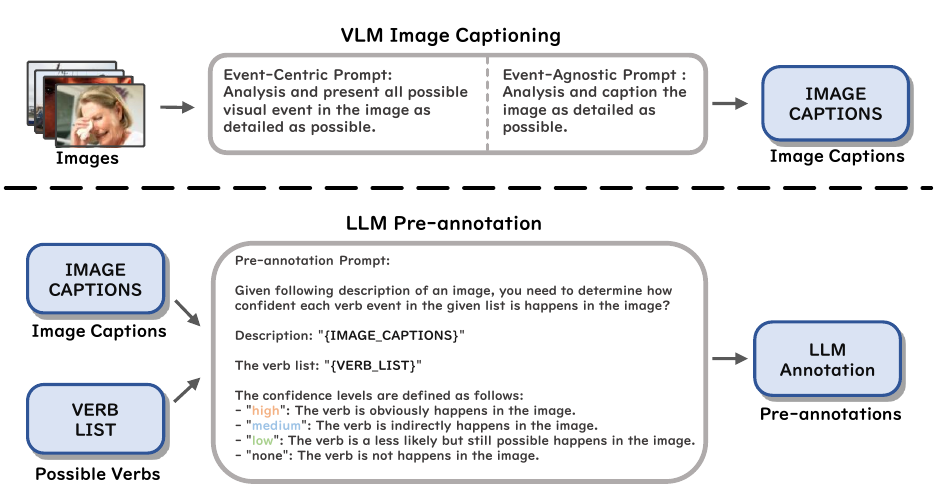}
\caption{
Two-stage annotation pipeline. (Top) VLM-based image caption extraction using dual prompting strategies (event-centric and event-agnostic) to ensure caption quality. (Bottom) LLM-driven pre-annotation where dynamically constructed prompts combine image captions with target categories to generate confidence scores for each class.
}
\label{fig:pre_annotation}
\end{figure}

As illustrated in Fig.~\ref{fig:pre_annotation}, our method first generates image captions using both event-centric and event-agnostic prompting strategies via Qwen2.5-VL~\cite{bai2025qwen25vltechnicalreport} and DeepSeek-VL2~\cite{wu2024deepseekvl2mixtureofexpertsvisionlanguagemodels}, preserving fine-grained visual and event information. The caption results from each model are then concatenated and fed into DeepSeek-V3~\cite{deepseekai2025deepseekv3technicalreport}, which predicts relevance levels for the top-100 candidate classes given by a single-positive-label-trained verb classifier. The model assigns relevance at four granularity levels (``high", ``medium", ``low", ``none") to facilitate manual annotation. When merging results from different models' captions, conflicting predictions are resolved by retaining the highest relevance level.

Our proposed pipeline achieves significant efficiency gains: The initial verb classifier reduces the average candidate set size from 504 to 100 classes per image while maintaining 98\% coverage of human-annotated labels (validated on our mini-benchmark). The subsequent VLM-LLM annotation further narrows this down to just 20.73 classes per image on average, while still covering 85\% of human-annotated labels.






\subsection{Human Annotation and Re-Annotation}
We recruited eight annotators to independently label the dataset based on pre-annotated data. To ensure a high recall, each image was presented with 100 candidate classes ranked by pre-annotated priority (``high", ``medium", ``low", ``none"), and annotators were instructed to mark all occurring visual events. To mitigate potential distributional biases between our annotations and the original dataset, we implemented a re-annotation step: For each class, annotators reviewed both the originally assigned images and the newly labeled ones, removing inconsistent labels to align the new annotations with the original label distribution. Finally, two experts performed cross-validation to ensure annotation quality.

\subsection{Dataset Characteristic}
We systematically expanded the annotations in the ImSitu test set (25,200 images), increasing the total number of annotations from 25,200 to 119,272 (averaging 4.74 labels per image), as detailed in Tab. \ref{tab:Characteristic}. As shown in Fig. \ref{fig:ablation_study_hyper_param}, the label frequencies follow a long-tailed distribution, with the top-20 most common categories highlighting the dataset's inherent multi-label nature.

\begin{table}[htbp]
  \centering
\footnotesize
  \renewcommand{\arraystretch}{1.0} 
  \caption{Dataset Characteristic}
  \label{tab:Characteristic}
  \begin{tabular}{lc}
    \toprule
    \textbf{\textit{Attribute}}& \textbf{\textit{Value}} \\
    \midrule
    Total Images & 25,200 \\
    Total Unique Labels & 504 \\
    Total Label Occurrences & 119,372 \\
    Average Labels per Image & 4.74 \\
    Median Labels per Image & 4 \\
    Min Labels per Image & 1 \\
    Max Labels per Image & 21 \\
    \bottomrule
  \end{tabular}
\end{table}

\begin{figure}
\vspace{-2mm}
  \centering
  \begin{subfigure}{0.49\linewidth}
    \includegraphics[width=1.05\textwidth]{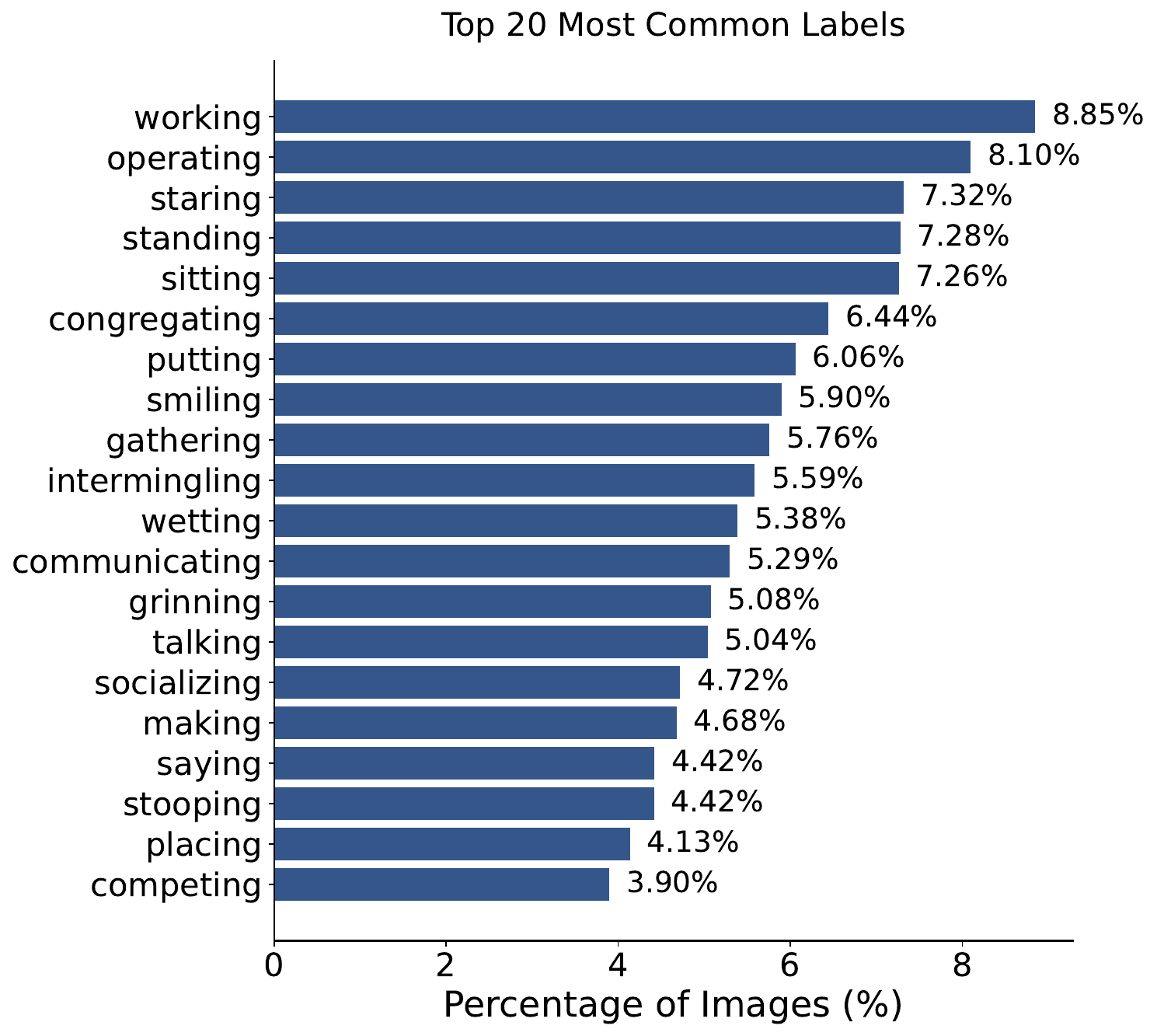}
    \label{fig:top_20_labels_horizontal}
  \end{subfigure}
  \hfill
  \begin{subfigure}{0.49\linewidth}
    \includegraphics[width=1.05\textwidth]{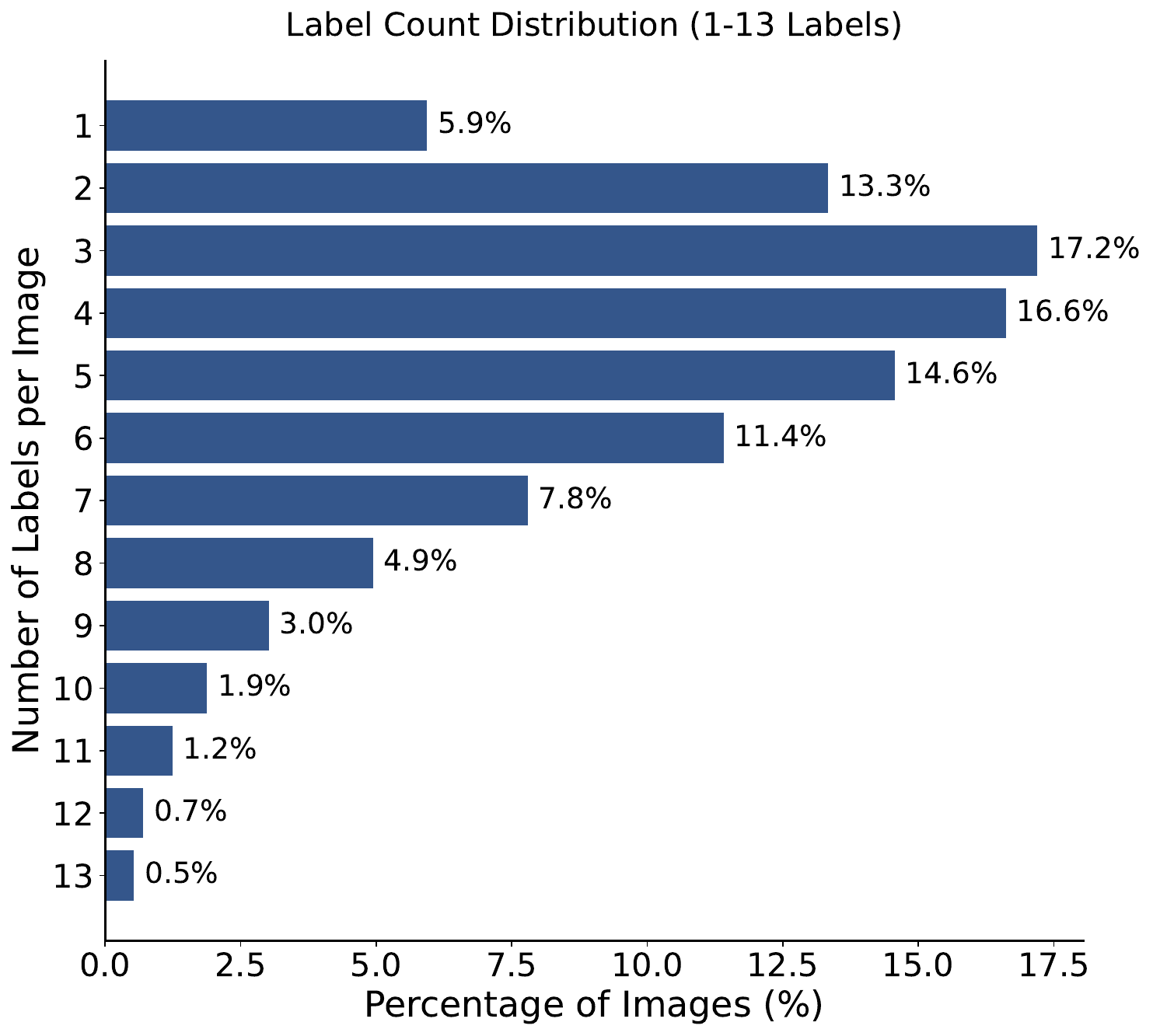}
    \label{fig:label_count_distribution_horizontal}
  \end{subfigure}
  \caption{
    Top-20 most common labels (left) and per-image label count distribution (right). 
    The left figure illustrates that there are a large number of categories where other classes may have ambiguous issues, further demonstrating that ambiguous problems are very common in the dataset.
    The right figure reveals that each image is generally associated with 2–6 labels, indicating a multi-label distribution. This label distribution further supports the idea that verb classification can be formulated as an SPMLL problem.
    }
  \label{fig:dataset_stat}
\end{figure}

\subsection{Benchmark on Verb Classification Literature}
To evaluate the model’s efficacy in multi-label classification within the existing literature, we replicate the results of previous verb classification models using publicly available code. We evaluate them using the multi-label Mean Average Precision (MAP) metric on our newly proposed mini-benchmark. We also include their conventional Top-1 and Top-5 accuracy results for comparison. The results are presented in the right part of Tab.~\ref{tab:label_static_and_SR_benchmark}, revealing a coherent improvement in both metrics. Notably, the MAP performance has shown a more pronounced upward trajectory in recent years, suggesting an enhanced capability of contemporary models in the SPMLL setting, especially for CLIP-based models (demonstrating a 10\% performance improvement). We attribute this enhancement to the inherently well-trained embedding distribution of the CLIP model.

\begin{table}[htbp]
  \centering
  \caption{Statistics of selected labels’ accuracy on original dataset (left) and benchmark on recent verb classification models (right)}
  \label{tab:label_static_and_SR_benchmark}   
    \begin{subtable}[t]{0.35\linewidth}
      \centering
      \footnotesize
      \setlength{\tabcolsep}{0.8pt} 
      \renewcommand{\arraystretch}{0.88} 
      \begin{tabular}{ccc}
        \toprule
        \textbf{\textit{Classes}}& \textbf{\textit{Top-1 Acc}} \\
        \midrule
        launching & 88.0\%\\
        climbing & 68.0\% \\
        swimming & 64.0\%\\
        aiming & 61.0\% \\
        crying & 52.0\%\\
        drawing & 42.0\%\\
        unloading & 42.0\%\\
        packaging & 40.0\%\\
        weeping & 34.0\%\\
        wetting & 0.0\%\\
      \bottomrule
      \end{tabular}
      \label{tab:label_anno_and_perf}
    \end{subtable}
    \hfill
    \begin{subtable}[t]{0.64\linewidth}
      \centering
      \footnotesize
      \setlength{\tabcolsep}{0.8pt} 
      \renewcommand{\arraystretch}{1.38} 
        \begin{tabular}{lccc}
            \toprule
            \textbf{\textit{Method}} & \textbf{\textit{Top-1 Acc}} & \textbf{\textit{Top-5 Acc}} & \textbf{\textit{MAP}} \\
            \midrule
            CRF {\tiny (CVPR 16')} & 33.2\% & 60.0\% & 34.8\% \\
            RE-VGG {\tiny (CVPR 20')}& 38.4\% & 65.0\% & 37.4\% \\
            JSL {\tiny (ECCV 20')}& 39.9\% & 67.6\% & 42.5\% \\
            GSRTR {\tiny (BMVC 21')}& 40.6\% & 69.8\% & 43.6\% \\
            CoFormer {\tiny (CVPR 22')}& 44.4\% & 72.9\% & 47.7\% \\
            ClipSitu {\tiny (WACV 24')}& 48.6\% & 78.2\% & 53.8\% \\
            \bottomrule
        \end{tabular}
        \label{tab:sr methods}   
    \end{subtable}
\end{table}
\section{Method}
\label{sec:method}


As discussed in Sec.~\ref{sec:problem formulation}, SPMLL aims to minimize the gap between $R_{\textrm{full}}(f_{\theta})$ and $R_{\textrm{partial}}$. This section introduces our proposed method for addressing this gap.

\subsection{Encoding Label Correlations via GCN}
By forming the verb classification task into the multi-label framework, the potential label space for model classification increases exponentially from $L$ to $2^L$. To handle such expansion, we propose to mine the co-occurrence between classes, which can be effectively represented using a label correlation matrix. However, the lack of ground-truth multi-label annotations makes distilling label correlation from the dataset problematic. Fortunately, in verb classification, class semantic similarity can be used as auxiliary information to partially reflect label correlation.

\begin{figure}[htbp]
\centering
\includegraphics[width=0.49\textwidth]{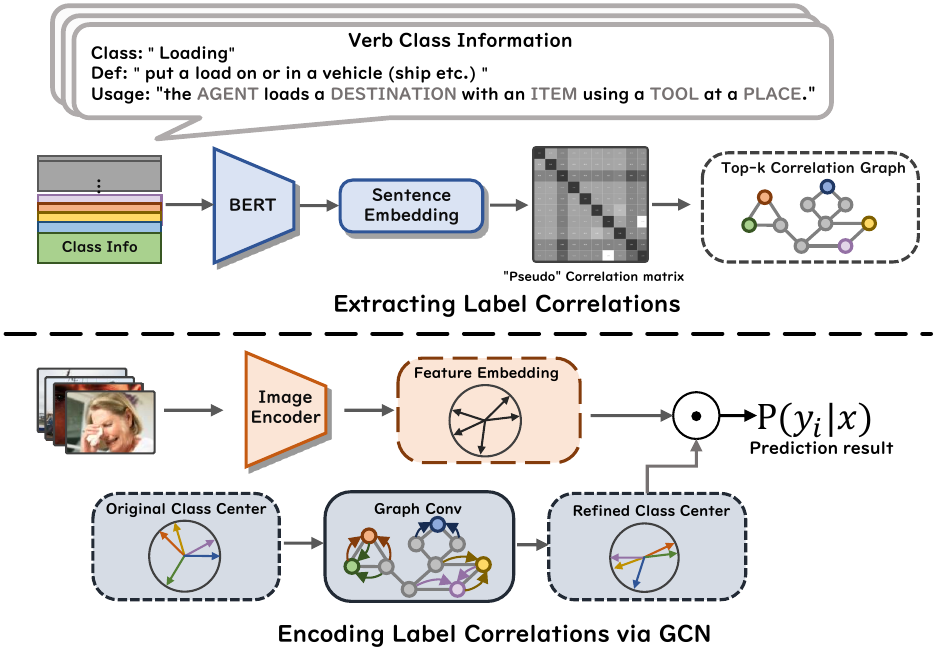}
\caption{
We use semantic relations to regularize the relations between categories in verb classification. Specifically, we first compute the semantic similarity between categories and use such similarity matrix as a proxy for label correlation to construct a sparse label correlation graph, which is then used in a multi-layer GCN to constrain the relations between classes systematically.}
\label{fig:model}
\end{figure}

Leveraging the above insights, we propose the Graph Enhanced Verb MLP (GE-VerbMLP) model. Specifically, we first compute the semantic similarity matrix between each class' definition as a ``pseudo" label correlation prior. Then, we change the linear dot product of the last layer to cosine similarity to describe the relationship between each category. Finally, we utilize the similarity matrix extracted above as a proxy for the correlation matrix and explicitly constrain the relationship between verb classes in the model through a graph convolutional  network. The overall model architecture is shown in Fig. ~\ref{fig:model}.

\subsubsection{Constructing Label Correlations}

Given that the verb classes of SR are derived from FrameNet \cite{fillmore2002framenet}, we calculate the semantic similarity between categories in FrameNet as a proxy of label correlation. Concretely, we compile the verb information in FrameNet in terms of the verb class name, verb definition, and usage frame. With the above keywords, we can form a sentence $s_i$ for each verb class and take BERT~\cite{reimers2019sentence} to obtain the semantic embedding of the sentence $c_i = \textrm{BERT}(s_i)$. Subsequently, the semantic similarity prior $A_{n\times n}$ between verb classes can be calculated as follows:

\begin{equation}
a_{ij} = \textrm{sim}(c_i, c_j) = \frac{c_i^T c_j}{\|c_i\|_2 \|c_j\|_2},
\end{equation}
where $\textrm{sim}(\cdot, \cdot)$ is the cosine similarity between two vectors.

In order to integrate the information of the most similar classes into the graph model, we sparsify the similarity matrix with $K$ Nearest Neighbor strategy. Specifically, for each class, we set all classes except the most similar $K$ classes to $0$, thus obtaining a sparse matrix $\bar{A}_{n \times n}$:

\begin{equation}
\bar{a}_{ij} = \left\{\begin{array}{ll}a_{ij},&\text{if}\ j\in\operatorname{top}K(\boldsymbol{a}_i),\\0,&\text{if}\ j\notin\operatorname{top}K(\boldsymbol{a}_i).\end{array}\right.
\end{equation}

Furthermore, we smooth the similarity matrix to improve the stability of the learning process, where the smoothed similarity matrix $\hat{A}$ is calculated as follows
\begin{equation}
\label{eqn:similarity-matrix}
\hat{a}_{ij}=\left\{\begin{array}{ll} s \frac{\bar{a}_{ij}}{\sum_{j' \neq j}^n \bar{a}_{ij'}} ,&\textrm{ if } i \neq j, \\1 - s,&\textrm{ if } i=j, \end{array}\right.
\end{equation}
where $s$ is a hyper-parameter, which limits the sum of the weights of all nodes to $s$. Finally, the label correlation is encoded in the sparse structured graph $\hat{A}$, which provides a rich structured semantic prior for our classification model.

\subsubsection{Extracting Image Visual Embedding from CLIP}
We follow the implementation of ClipSitu Verb MLP~\cite{roy2024clipsitu} for image embedding extraction. Given an input image $x_i$, its visual embedding $e_{i}$ is extracted by a weight-frozen CLIP image encoder follows an multi-layer perceptron: 
\begin{equation}
e_{i} = f_{\textrm{MLP}}(f_{\textrm{CLIP}}(x_{i})),
\end{equation}
where $f_{\textrm{MLP}}$ contains $l$ trainable linear layers of a fixed dimension with ReLU activation to get embedding for the specific verb classification task. Finally, we learn the projection vector $e_i$ of each image in the class semantic space.



\subsubsection{Classification with Graph Convolutional Network } 
Given the class center vectors $\{c_0,c_1,\cdots,c_L\}$ of $L$ classes, we use $J$ graph convolutional network layers to gradually fine-tune the class center vectors, where the output of the $j$-th layer of GCN is the input of the next layer of GCN
\begin{equation}
C_{j+1} = \rho(\hat{A}C_j W_j),\quad j = 1,2,\cdots, J.
\end{equation}
Here $\hat{A}$ is the similarity matrix as shown in Eqn. (\ref{eqn:similarity-matrix}), $W_j$ is the learnable parameter of the $j$-th layer GCN, and $\rho(\cdot)$ is the nonlinear activation function. 
Initially, we denote
\begin{equation}
        C_1 = 
	\begin{bmatrix} 
            c_0 & c_1 & \cdots & c_L 
	\end{bmatrix},
\end{equation}
then, through $J$ iterations, we get $C_{J + 1}$. 

The intuition here is that by utilizing the sparsely connected attention graph $A$ as a prior, for each class center vector (e.g., ``wetting,"), the model aggregates the class center features of the $k$ most similar verbs (e.g., ``swimming," ``watering") in an iterative manner through graph convolution as a new representation. After performing multiple graph convolution layers, the distances between similar verb classes will be closer, meaning that overlapping classes are more likely to be classified simultaneously.

Subsequently, we use residual connection to retain the information of the original class center vector
$\hat{C} = C_1 + C_{J + 1}$. Similarly, given a sample $x_i$, the posterior probability of each class under the class vector matrix $\hat{C}$ is
\begin{equation}
p(j|x_{i}) = \sigma(\textrm{sim}(e_{i}, \hat{c}_j)/\tau),
\end{equation}
where $\hat{c}_j$ is the $j$-th column of $\hat{C}$, and $\tau$ represents the temperature used to adjust the sensitivity of the output to similarity. In our experiments, $\tau$ is empirically set to 10, and we randomly initialize the class center $c_j$ of the $j$th class.

\subsection{Incoporating Adversarial Training}

By leveraging GCN to encode label correlation information into the model, we can cluster the class center between overlapped classes. However, when dealing with classes that exhibit subtle differences, it becomes crucial to employ effective methods for their differentiation. Motivated by such insight, we further integrate adversarial training into our approach to facilitate learning smoother classification boundaries.
Specifically, we take the Fast Gradient Sign Method (FGSM) \cite{goodfellow2014explaining} and Projected Gradient Descent (PGD) \cite{madry2017towards}  to generate hard examples during training.

FGSM creates a perturbation $\delta$ by linearizing the loss function around the current value of $\theta$,
\begin{equation}
\delta_{\textrm{FGSM}} = \epsilon \text{sign}(\nabla_{x} L(f_{\theta}(x), z)),
\end{equation}
where $\epsilon$ is the learning step size and $\nabla$ represents the gradient. Hence, the adversarial sample generated by $x$ is
\begin{equation}
\label{eqn:fgsm}
x_{\textrm{FGSM}} = x + \delta_{\textrm{FGSM}}.
\end{equation}

Compared with the FGSM attack that uses only one gradient descent iteration, the PGD attack has multiple additional projection operations. Specifically, the adversarial examples generated by the PGD attack are shown below
\begin{equation}
x_{\textrm{PGD}} = \Pi_S(x + \delta_{\textrm{FGSM}}),
\end{equation}
where $\delta_{\textrm{FGSM}}$ is calculated by Eqn.~(\ref{eqn:fgsm}) and $\Pi_S(x)$ is the projection of $x$ onto the set $S$. The projection  ensures that the adversarial example always lies within the neighborhood $S$ centered on $x$.

\section{Experiments}
\label{sec:expriment}
In this section, we conduct experiments on both the imSitu dataset and the new benchmark. First, we detail the experimental setup, compare various methods with our proposed models on both datasets and discuss the results. Next, we perform ablation studies to evaluate the impact of integrating GCNs and adversarial training. Finally, we present an analysis of hyper-parameter sensitivity.

\subsection{Experimental Settings}

\subsubsection{Dataset Settings and Evaluation Metrics}
The imSitu dataset, the only publicly available benchmark in SR, includes 126,102 images of 504 classes. 
Following the setting in the literature~\cite{yatskar2016situation}, we present the top-1 and top-5 accuracy on the original imSitu dataset. At the same time, on our proposed multi-label benchmark, we compare the multi-label classification ability of different models by the criterion of mean average precision (MAP). 
The Top-K accuracy reflects the model’s alignment with annotators’ primary perceptions but does not assess its ability to handle ambiguity. In contrast, MAP provides a more robust evaluation metric by considering all possible labels but may diverge from human initial intuition. We present both metrics together, as an ideal model should excel in both aspects.

\begin{table}[htbp]
  \centering
  \footnotesize
  \caption{Performance benchmark of our method and other methods in different settings: Top-1 and Top-5 acc are used to measure the performance of multi-class classification, while MAP is adopted to measure the performance in multi-label setting.}
  \label{tab:1024_benchmark}
  \begin{tabular}{lcccc}
    \toprule
    \textbf{\textit{Task Setting}}&\textbf{\textit{Method}}& \textbf{\textit{Top-1 Acc}}& \textbf{\textit{Top-5 Acc}}& \textbf{\textit{MAP}}\\
    \midrule
    Multi-class & CE (CLIPSitu) & 48.6\%&  78.2\% & 53.8\% \\
    \hline
    Multi-label & BCE & 48.4\% &  77.7\% & 55.1\% \\
    Multi-label & Focal & \textbf{48.7}\% &  \textbf{77.8\%} & \textbf{55.1\%} \\
    \hline
    SPMLL & SMILE\_si & 48.1\% & 77.6\% & 48.7\% \\
    SPMLL & SMILE & 47.7\% & 77.6\% & 49.5\% \\
    SPMLL & EPR & 42.1\% & 75.9\% & 50.7\%\\
    SPMLL & WAN & 42.2\% & 74.3\% & 52.5\% \\
    SPMLL & SPLC & \textbf{48.8}\%&  77.7\% & 54.5\% \\
    SPMLL & EM & 40.3\% & 75.0\% & 54.6\% \\
    SPMLL & EM-APL & 40.4\% & 75.2\% & 54.6\% \\
    SPMLL & BCE-LS & 48.4\% & 77.9\% & 55.5\% \\
    SPMLL & SCPNet & 48.3\% & 78.0\% & 55.8\% \\
    SPMLL & ROLE & 47.4\% & 77.8\% & 56.2\% \\
    SPMLL & Ours & 48.3\%&  \textbf{78.1}\% & \textbf{57.0}\% \\
    \bottomrule
  \end{tabular}
\end{table}

\subsubsection{Implementation Details}
For the model backbone, we adhere to the implementation of the state-of-the-art ClipSitu Verb MLP model~\cite{roy2024clipsitu}. Specifically, to ensure fair comparison, we employ the pre-trained ViT-B32 model as the frozen backbone and incorporate two layers of MLP with a hidden dimension of 1,024 and the ReLU activation function as feature encoder for all compared model. All models are trained using the Adam optimizer with a learning rate of $2 \times 10^{-4}$ and an exponential learning rate scheduler with a discount factor of $0.9$. In our proposed method, we utilize the scaled cosine similarity between image embeddings and class center vectors as logits. During training, the parameters of the multi-layer MLP, class centers, and GCN are updated.

\subsubsection{Benchmark Methods and Task Settings}
This section presents the comparison methods across multi-class, multi-label, and single positive multi-label learning.

For multi-class setting, we employ the standard cross-entropy loss, which is widely adopted in the literature. For multi-label setting, we evaluate two approaches: Binary Cross-Entropy (BCE) loss and Focal loss~\cite{lin2017focal}.



Then, we adapte current methods in the SPMLL literature on the verb classification task, and present their benchmark result.
We first implement several classical approaches from the literature~\cite{cole2021multi}, including: 1). WAN loss, which reduces the weights of all negative labels, 2). BCE-LS loss, which has demonstrated effectiveness for noisy label problems, 3). EPR loss, which focuses on regulating the expected number of positive labels, and 4). the ROLE method, which extends EPR loss with a label estimator.
Additionally, we incorporate more recent SPMLL methods for comparison, including novel loss functions such as 5). EM loss~\cite{zhou2022acknowledging}, 6). alternative model architectures like the SCPNet~\cite{ding2023exploring}, and pseudo-labeling methods such as 7). the SPLC~\cite{zhang2021simple}, 8). SMILE~\cite{xu2022one}, and  9). EM-APL loss~\cite{zhou2022acknowledging}.
Finally, we report the performance of our proposed method that is specially designed for the ambiguity problem in the verb classification task.

\subsection{Experimental Results}
\subsubsection{Comparison Results}
Tab.~\ref{tab:1024_benchmark} presents the performance benchmarks of all the aforementioned methods under both the original Top-K accuracy metric and the new multi-label MAP metric.

Compared to baseline methods, the multi-label approaches achieve immediate performance improvement in the MAP metric. Notably, such improvement does not result in a significant drop in the Top-K accuracy, suggesting potential inadequacies in the existing multi-class training schemes.

Comparing the SPMLL-based methods with the baseline methods, we observe varying effectiveness across different SPMLL methods on multi-label evaluation benchmark. 

For EPR, WAN, and EM variants, their limited Top-1/Top-5 accuracies indicate that modifying the loss function to handle negative labels may inadvertently compromise classification performance on the ImSitu dataset. As for the SMILE variants, despite their theoretically well-founded VAE-based framework, their effectiveness remains limited—likely due to the inherent semantic ambiguity in our dataset. Additionally, constructing a full training-set adjacency matrix incurs prohibitively high computational costs.
In the case of SPLC, directly applying pseudo-labels to high-confidence predictions fails to improve MAP. Meanwhile, SCPNet employs a GCN module similar to our approach, but its self-supervised strong-weak augmentation yields only marginal MAP gains. Further, the ROLE method demonstrates that integrating label estimators stabilizes training and leads to better performance, albeit at the cost of reduced Top-1 accuracy.


Finally, it is noteworthy that our proposed method achieves  substantial improvement in the multi-label MAP metric while maintaining the original accuracy. These results highlight the effectiveness of our approach in boosting multi-label classification performance without sacrificing the model’s overall accuracy.

\begin{table}[htbp]
  \centering
  \footnotesize
  \caption{Ablation studies on GCN \& adversarial training}
  \label{tab:ablation_gcn_adv}
  \begin{tabular}{ccccc}
    \toprule
    \textbf{\textit{With Adv}}& \textbf{\textit{With GCN}}& \textbf{\textit{Top-1 Acc}}& \textbf{\textit{Top-5 Acc}}& \textbf{\textit{MAP}} \\
    \midrule
    \XSolidBrush & \XSolidBrush & \textbf{48.6}\%&  77.9\% & 55.1\% \\
    \XSolidBrush & \CheckmarkBold & 48.3\%&  77.9\% & 55.9\% \\
    \CheckmarkBold & \XSolidBrush & 48.3\%&  \textbf{78.2\%} & 56.6\% \\
    \CheckmarkBold & \CheckmarkBold & 48.3\%&  78.1\% & \textbf{57.0\%} \\
    \bottomrule
  \end{tabular}
\end{table}

\subsubsection{Ablation Studies}

To better understand our proposed model, we examine the effectiveness of GCN and adversarial training in Tab.~\ref{tab:ablation_gcn_adv}. As observed, adversarial training and the addition of GCN indeed enhance the multi-label classification ability of the model. Their combination is observed to generate a significant gain in multi-label classification.


Furthermore, for adversarial training, we highlight the effects of FGSM and PGD on the model in Tab.~\ref{tab:comp_adv}. We find that both FGSM and PGD do not hurt the top-1 and top-5 accuracy, and more accurate gradients, as estimated by the PGD method, lead to higher MAP performance gain. In practice, we choose PGD as our adversarial training method.

\begin{table}[htbp]
  \centering
\footnotesize
  \caption{Ablation study on adversarial training methods}
  \label{tab:comp_adv}
  \begin{tabular}{lccc}
    \toprule
    \textbf{\textit{Method}}& \textbf{\textit{Top-1 Acc}}& \textbf{\textit{Top-5 Acc}}& \textbf{\textit{MAP}} \\
    \midrule
    GE+VerbMLP & 48.3\%&  77.9\% & 55.9\% \\
    GE+VerbMLP+FGSM & \textbf{48.5\%}&  \textbf{78.2}\% & 56.3\% \\
    GE+VerbMLP+PGD & 48.3\%&  78.1\% & \textbf{57.0\%} \\
    \bottomrule
  \end{tabular}
\end{table}

\subsubsection{Visualization of  Embedded Class Centers}
\begin{figure}[htbp]
\centering
\includegraphics[width=0.42\textwidth]{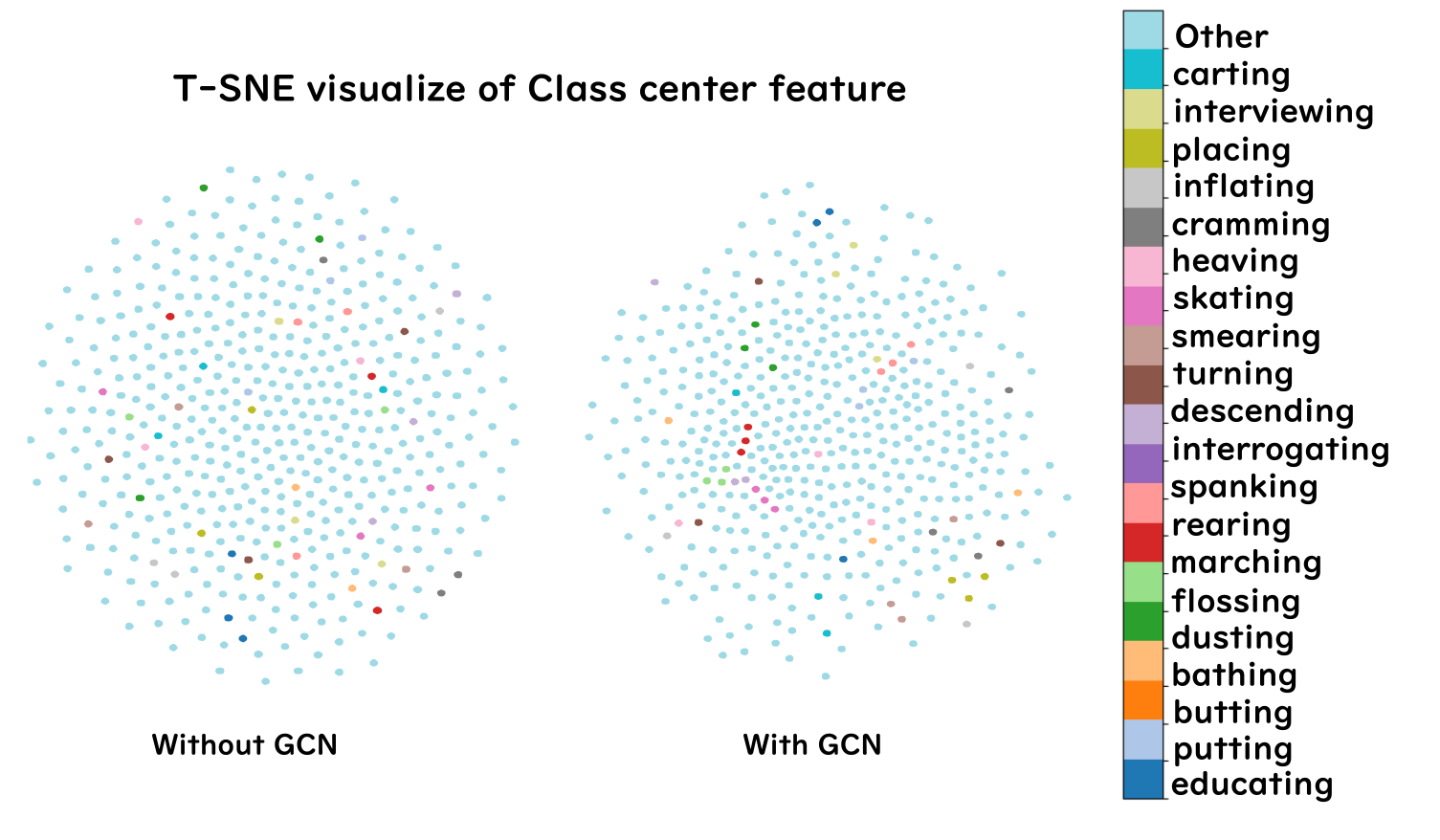}
\caption{T-SNE visualization of different class center vectors. Clearly, semantically similar classes are closer after GCN layers. Each point represents the center of a class, and points with the same color indicate that these classes are semantically similar (measured by $3$ nearest neighbors), where the color bar shows the anchor class used to calculate the three nearest neighbors.} 
\label{fig:gcn_vis}
\end{figure}

We take t-SNE \cite{van2008visualizing} to visually compare the relationship between class centers in the latent space during normal training and GCN refinement, as shown in Fig.~\ref{fig:gcn_vis}.
We can see that after applying GCN, semantically similar class centers are clustered more closely while still maintaining the distinguishability between different categories. 

\subsubsection{Analysis of GCN parameters $K$ and $J$}
\begin{figure}
  \centering
  \begin{subfigure}{0.49\linewidth}
    \includegraphics[width=1.05\textwidth]{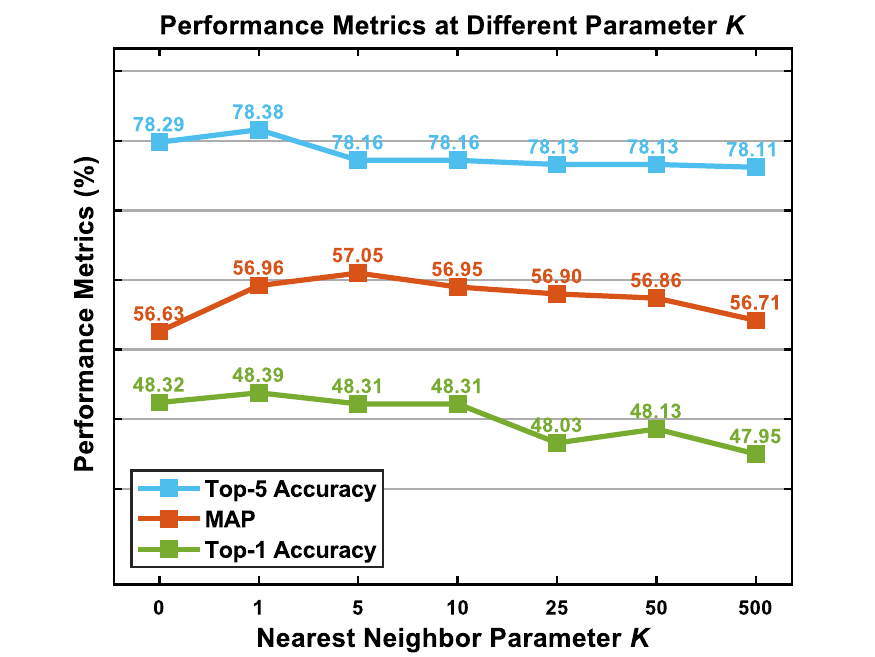}
    \label{fig:ablation_study_k}
  \end{subfigure}
  \hfill
  \begin{subfigure}{0.49\linewidth}
    \includegraphics[width=1.05\textwidth]{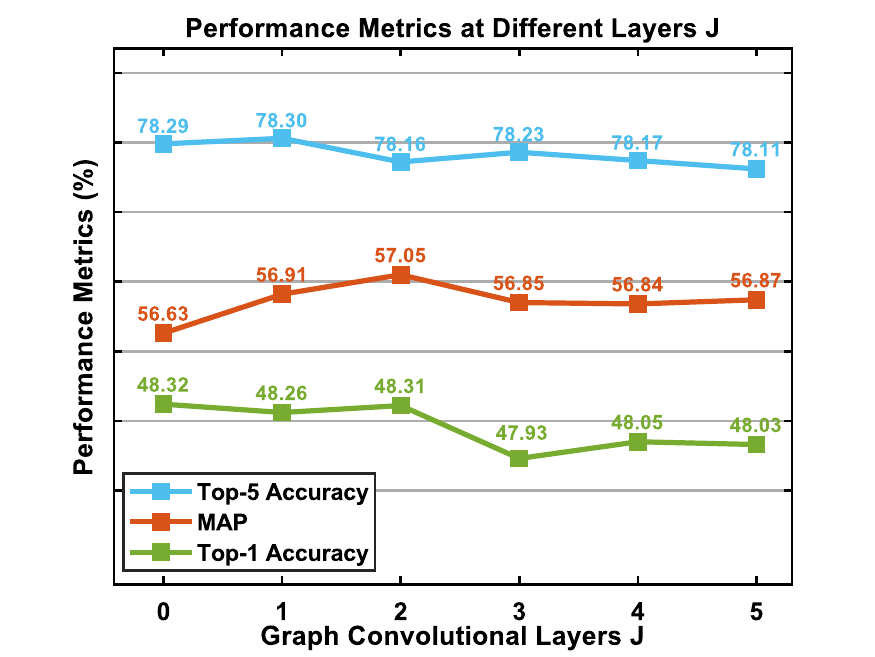}
    \label{fig:ablation_study_j}
  \end{subfigure}
  \caption{
    Analysis of hyper-parameters $K$ and $L$.
    }
  \label{fig:ablation_study_hyper_param}  
\end{figure}

In this section, we analyze the influence of parameters associated with our proposed GCN module in Fig.~\ref{fig:ablation_study_hyper_param}. On the left side of Fig.~\ref{fig:ablation_study_hyper_param}, we investigate the impact of adjusting the nearest neighbor number $K$ while constructing class correlation graph $\hat{A}$. Increasing the amount of $K$ results in a denser graph connection, leading to a more uniform distribution of class center vectors. It can be observed that a lower value of $K$ yields more effective results. Such observation is consistent with our findings in Fig.~\ref{fig:10_class_sep_heatmap} and Fig.~\ref{fig:dataset_stat}, indicating that classes tend to overlap with only a few specific classes. To the right side of Fig.~\ref{fig:ablation_study_hyper_param}, we explore the influence of varying the number of GCN layers $J$. As the number of layers increases, the receptive field expands, resulting in a smoother class center embedding. We observe a noticeable decrease in MAP beyond two GCN layers, which indicates that excessively smooth class centers may negatively impact performance. Moreover, it is noteworthy that the accuracy is not significantly sensitive to both $K$ and $J$, which underscores the robustness of our proposed method.

\section{Conclusion}
\label{sec:conclusion}

This paper presents a comprehensive study of the ambiguity problem of verb classification in context recognition (SR). Through extensive empirical analysis, we show that current single-label classification formulations fail to capture the inherent semantic overlap between verb categories, resulting in suboptimal performance and evaluation results. Our core insight is that verb classification should be fundamentally reformulated as a multi-label learning problem to better reflect visual event recognition's nature.


To address the practical challenge of obtaining full multi-label annotations, we propose to formulate verb classification as a single forward multi-label learning (SPMLL) problem—a new perspective in SR research. Our contributions include: 1) Deeply analyzing verb ambiguity through embeddings and manual annotations; 2) creating a large-scale multi-label evaluation benchmark to enable proper evaluation of SR models; and 3) developing GE-VerbMLP, which combines GNNs and adversarial training for robust performance. Experiments show GE-VerbMLP improves multi-label accuracy by over 3\% while maintaining competitive top-1/5 performance. These findings suggest that properly handling ambiguity leads to a more comprehensive understanding of context. Our benchmark and evaluation scheme provide researchers with the necessary tools to more accurately evaluate model performance in real-world scenarios where multiple verb interpretations may be valid.

Furthermore, our work opens up several promising research directions. First, the SPMLL framework can be extended to other components of SR, such as semantic role labeling. Second, more sophisticated label correlation learning methods can further improve the performance of multi-label classification. Finally, our results demonstrate that richer labeling schemes can capture valid interpretations in visual event recognition. We believe that these advances will enable SR systems to better match the human level of context understanding.

In summary, by identifying and resolving fundamental ambiguity in verb classification, our work provides theoretical insights and practical solutions for advancing context recognition research. The proposed methods and benchmarks lay the foundation for developing more powerful and nuanced visual understanding systems.


\newpage
{
    \small
    \bibliographystyle{IEEEtran}
    \bibliography{main}

\begin{thebibliography}{10}
\providecommand{\url}[1]{#1}
\csname url@samestyle\endcsname
\providecommand{\newblock}{\relax}
\providecommand{\bibinfo}[2]{#2}
\providecommand{\BIBentrySTDinterwordspacing}{\spaceskip=0pt\relax}
\providecommand{\BIBentryALTinterwordstretchfactor}{4}
\providecommand{\BIBentryALTinterwordspacing}{\spaceskip=\fontdimen2\font plus
\BIBentryALTinterwordstretchfactor\fontdimen3\font minus \fontdimen4\font\relax}
\providecommand{\BIBforeignlanguage}[2]{{%
\expandafter\ifx\csname l@#1\endcsname\relax
\typeout{** WARNING: IEEEtran.bst: No hyphenation pattern has been}%
\typeout{** loaded for the language `#1'. Using the pattern for}%
\typeout{** the default language instead.}%
\else
\language=\csname l@#1\endcsname
\fi
#2}}
\providecommand{\BIBdecl}{\relax}
\BIBdecl

\bibitem{cole2021multilabellearningsinglepositive}
\BIBentryALTinterwordspacing
E.~Cole, O.~M. Aodha, T.~Lorieul, P.~Perona, D.~Morris, and N.~Jojic, ``Multi-label learning from single positive labels,'' 2021. [Online]. Available: \url{https://arxiv.org/abs/2106.09708}
\BIBentrySTDinterwordspacing

\bibitem{qian2022survey}
Z.~Qian, K.~Huang, Q.-F. Wang, and X.-Y. Zhang, ``A survey of robust adversarial training in pattern recognition: Fundamental, theory, and methodologies,'' \emph{Pattern Recognition}, vol. 131, p. 108889, 2022.

\bibitem{goodfellow2014explaining}
I.~J. Goodfellow, J.~Shlens, and C.~Szegedy, ``Explaining and harnessing adversarial examples,'' \emph{arXiv preprint arXiv:1412.6572}, 2014.

\bibitem{madry2017towards}
A.~Madry, A.~Makelov, L.~Schmidt, D.~Tsipras, and A.~Vladu, ``Towards deep learning models resistant to adversarial attacks,'' \emph{arXiv preprint arXiv:1706.06083}, 2017.

\bibitem{goodfellow2014generative}
I.~Goodfellow, J.~Pouget-Abadie, M.~Mirza, B.~Xu, D.~Warde-Farley, S.~Ozair, A.~Courville, and Y.~Bengio, ``Generative adversarial nets,'' \emph{Advances in Neural Information Processing Systems}, vol.~27, 2014.

\bibitem{liu2019gandef}
G.~Liu, I.~Khalil, and A.~Khreishah, ``Gandef: A gan based adversarial training defense for neural network classifier,'' in \emph{ICT Systems Security and Privacy Protection: 34th IFIP TC 11 International Conference, SEC 2019, Lisbon, Portugal, June 25-27, 2019, Proceedings 34}.\hskip 1em plus 0.5em minus 0.4em\relax Springer, 2019, pp. 19--32.

\bibitem{ho2020denoising}
J.~Ho, A.~Jain, and P.~Abbeel, ``Denoising diffusion probabilistic models,'' \emph{Advances in Neural Information Processing Systems}, vol.~33, pp. 6840--6851, 2020.

\bibitem{lv2019weakly}
J.~Lv, N.~Xu, R.~Zheng, and X.~Geng, ``Weakly supervised multi-label learning via label enhancement.'' in \emph{IJCAI}, 2019, pp. 3101--3107.

\bibitem{cole2021multi}
E.~Cole, O.~Mac~Aodha, T.~Lorieul, P.~Perona, D.~Morris, and N.~Jojic, ``Multi-label learning from single positive labels,'' in \emph{Proceedings of the IEEE/CVF Conference on Computer Vision and Pattern Recognition}, 2021, pp. 933--942.

\bibitem{muller2019does}
R.~M{\"u}ller, S.~Kornblith, and G.~E. Hinton, ``When does label smoothing help?'' \emph{Advances in Neural Information Processing Systems}, vol.~32, 2019.

\bibitem{zhang2021simple}
Y.~Zhang, Y.~Cheng, X.~Huang, F.~Wen, R.~Feng, Y.~Li, and Y.~Guo, ``Simple and robust loss design for multi-label learning with missing labels,'' \emph{arXiv preprint arXiv:2112.07368}, 2021.

\bibitem{kim2022large}
Y.~Kim, J.~M. Kim, Z.~Akata, and J.~Lee, ``Large loss matters in weakly supervised multi-label classification,'' in \emph{Proceedings of the IEEE/CVF Conference on Computer Vision and Pattern Recognition}, 2022, pp. 14\,156--14\,165.

\bibitem{kim2023bridging}
Y.~Kim, J.~M. Kim, J.~Jeong, C.~Schmid, Z.~Akata, and J.~Lee, ``Bridging the gap between model explanations in partially annotated multi-label classification,'' in \emph{Proceedings of the IEEE/CVF Conference on Computer Vision and Pattern Recognition}, 2023, pp. 3408--3417.

\bibitem{ding2023exploring}
Z.~Ding, A.~Wang, H.~Chen, Q.~Zhang, P.~Liu, Y.~Bao, W.~Yan, and J.~Han, ``Exploring structured semantic prior for multi label recognition with incomplete labels,'' in \emph{Proceedings of the IEEE/CVF Conference on Computer Vision and Pattern Recognition}, 2023, pp. 3398--3407.

\bibitem{liu2023revisiting}
B.~Liu, N.~Xu, J.~Lv, and X.~Geng, ``Revisiting pseudo-label for single-positive multi-label learning,'' in \emph{International Conference on Machine Learning}.\hskip 1em plus 0.5em minus 0.4em\relax PMLR, 2023, pp. 22\,249--22\,265.

\bibitem{wang2023hierarchical}
A.~Wang, H.~Chen, Z.~Lin, Z.~Ding, P.~Liu, Y.~Bao, W.~Yan, and G.~Ding, ``Hierarchical prompt learning using clip for multi-label classification with single positive labels,'' in \emph{Proceedings of the 31st ACM International Conference on Multimedia}, 2023, pp. 5594--5604.

\bibitem{roy2024clipsitu}
D.~Roy, D.~Verma, and B.~Fernando, ``Clipsitu: Effectively leveraging clip for conditional predictions in situation recognition,'' in \emph{Proceedings of the IEEE/CVF Winter Conference on Applications of Computer Vision}, 2024, pp. 444--453.

\bibitem{li2022clip}
M.~Li, R.~Xu, S.~Wang, L.~Zhou, X.~Lin, C.~Zhu, M.~Zeng, H.~Ji, and S.-F. Chang, ``Clip-event: Connecting text and images with event structures,'' in \emph{2022 IEEE/CVF Conference on Computer Vision and Pattern Recognition (CVPR)}.\hskip 1em plus 0.5em minus 0.4em\relax IEEE, 2022, pp. 16\,399--16\,408.

\bibitem{mallya2017recurrent}
A.~Mallya and S.~Lazebnik, ``Recurrent models for situation recognition,'' in \emph{Proceedings of the IEEE International Conference on Computer Vision}, 2017, pp. 455--463.

\bibitem{li2017situation}
R.~Li, M.~Tapaswi, R.~Liao, J.~Jia, R.~Urtasun, and S.~Fidler, ``Situation recognition with graph neural networks,'' in \emph{Proceedings of the IEEE International Conference on Computer Vision}, 2017, pp. 4173--4182.

\bibitem{suhail2019mixture}
M.~Suhail and L.~Sigal, ``Mixture-kernel graph attention network for situation recognition,'' in \emph{Proceedings of the IEEE/CVF International Conference on Computer Vision}, 2019, pp. 10\,363--10\,372.

\bibitem{cooray2020attention}
T.~Cooray, N.-M. Cheung, and W.~Lu, ``Attention-based context aware reasoning for situation recognition,'' in \emph{Proceedings of the IEEE/CVF Conference on Computer Vision and Pattern Recognition}, 2020, pp. 4736--4745.

\bibitem{cho2022collaborative}
J.~Cho, Y.~Yoon, and S.~Kwak, ``Collaborative transformers for grounded situation recognition,'' in \emph{Proceedings of the IEEE/CVF Conference on Computer Vision and Pattern Recognition}, 2022, pp. 19\,659--19\,668.

\bibitem{sanders2022ambiguous}
K.~Sanders, R.~Kriz, A.~Liu, and B.~Van~Durme, ``Ambiguous images with human judgments for robust visual event classification,'' \emph{Advances in Neural Information Processing Systems}, vol.~35, pp. 2637--2650, 2022.

\bibitem{yatskar2016situation}
M.~Yatskar, L.~Zettlemoyer, and A.~Farhadi, ``Situation recognition: Visual semantic role labeling for image understanding,'' in \emph{Proceedings of the IEEE Conference on Computer Vision and Pattern Recognition}, 2016, pp. 5534--5542.

\bibitem{pratt2020grounded}
S.~Pratt, M.~Yatskar, L.~Weihs, A.~Farhadi, and A.~Kembhavi, ``Grounded situation recognition,'' in \emph{Computer Vision--ECCV 2020: 16th European Conference, Glasgow, UK, August 23--28, 2020, Proceedings, Part IV 16}.\hskip 1em plus 0.5em minus 0.4em\relax Springer, 2020, pp. 314--332.

\bibitem{cho2021grounded}
J.~Cho, Y.~Yoon, H.~Lee, and S.~Kwak, ``Grounded situation recognition with transformers,'' \emph{arXiv preprint arXiv:2111.10135}, 2021.

\bibitem{van2008visualizing}
L.~Van~der Maaten and G.~Hinton, ``Visualizing data using t-sne.'' \emph{Journal of Machine Learning Research}, vol.~9, no.~11, 2008.

\bibitem{kipf2016semi}
T.~N. Kipf and M.~Welling, ``Semi-supervised classification with graph convolutional networks,'' \emph{arXiv preprint arXiv:1609.02907}, 2016.

\bibitem{radford2021learning}
A.~Radford, J.~W. Kim, C.~Hallacy, A.~Ramesh, G.~Goh, S.~Agarwal, G.~Sastry, A.~Askell, P.~Mishkin, J.~Clark \emph{et~al.}, ``Learning transferable visual models from natural language supervision,'' in \emph{International Conference on Machine Learning}.\hskip 1em plus 0.5em minus 0.4em\relax PMLR, 2021, pp. 8748--8763.

\bibitem{fillmore2002framenet}
C.~J. Fillmore, C.~F. Baker, and H.~Sato, ``The framenet database and software tools.'' in \emph{LREC}.\hskip 1em plus 0.5em minus 0.4em\relax Citeseer, 2002.

\bibitem{reimers2019sentence}
N.~Reimers and I.~Gurevych, ``Sentence-bert: Sentence embeddings using siamese bert-networks,'' \emph{arXiv preprint arXiv:1908.10084}, 2019.

\bibitem{lin2017focal}
T.-Y. Lin, P.~Goyal, R.~Girshick, K.~He, and P.~Doll{\'a}r, ``Focal loss for dense object detection,'' in \emph{Proceedings of the IEEE International Conference on Computer Vision}, 2017, pp. 2980--2988.

\bibitem{zhou2016learning}
B.~Zhou, A.~Khosla, A.~Lapedriza, A.~Oliva, and A.~Torralba, ``Learning deep features for discriminative localization,'' in \emph{Proceedings of the IEEE Conference on Computer Vision and Pattern Recognition}, 2016, pp. 2921--2929.

\bibitem{cordts2016cityscapes}
M.~Cordts, M.~Omran, S.~Ramos, T.~Rehfeld, M.~Enzweiler, R.~Benenson, U.~Franke, S.~Roth, and B.~Schiele, ``The cityscapes dataset for semantic urban scene understanding,'' in \emph{Proceedings of the IEEE Conference on Computer Vision and Pattern Recognition}, 2016, pp. 3213--3223.

\bibitem{caba2015activitynet}
F.~Caba~Heilbron, V.~Escorcia, B.~Ghanem, and J.~Carlos~Niebles, ``Activitynet: A large-scale video benchmark for human activity understanding,'' in \emph{Proceedings of the IEEE Conference on Computer Vision and Pattern Recognition}, 2015, pp. 961--970.

\bibitem{gu2019survey}
Y.~Gu, Y.~Wang, and Y.~Li, ``A survey on deep learning-driven remote sensing image scene understanding: Scene classification, scene retrieval and scene-guided object detection,'' \emph{Applied Sciences}, vol.~9, no.~10, p. 2110, 2019.

\bibitem{buxton2003learning}
H.~Buxton, ``Learning and understanding dynamic scene activity: a review,'' \emph{Image and Vision Computing}, vol.~21, no.~1, pp. 125--136, 2003.

\bibitem{zhou2022acknowledging}
D.~Zhou, P.~Chen, Q.~Wang, G.~Chen, and P.-A. Heng, ``Acknowledging the unknown for multi-label learning with single positive labels,'' in \emph{European Conference on Computer Vision}.\hskip 1em plus 0.5em minus 0.4em\relax Springer, 2022, pp. 423--440.

\bibitem{xu2022one}
N.~Xu, C.~Qiao, J.~Lv, X.~Geng, and M.-L. Zhang, ``One positive label is sufficient: Single-positive multi-label learning with label enhancement,'' \emph{Advances in Neural Information Processing Systems}, vol.~35, pp. 21\,765--21\,776, 2022.

\bibitem{bai2025qwen25vltechnicalreport}
\BIBentryALTinterwordspacing
S.~Bai and K.~C. et~al., ``Qwen2.5-vl technical report,'' 2025. [Online]. Available: \url{https://arxiv.org/abs/2502.13923}
\BIBentrySTDinterwordspacing

\bibitem{wu2024deepseekvl2mixtureofexpertsvisionlanguagemodels}
\BIBentryALTinterwordspacing
Z.~Wu and X.~C. et~al., ``Deepseek-vl2: Mixture-of-experts vision-language models for advanced multimodal understanding,'' 2024. [Online]. Available: \url{https://arxiv.org/abs/2412.10302}
\BIBentrySTDinterwordspacing

\bibitem{deepseekai2025deepseekv3technicalreport}
\BIBentryALTinterwordspacing
DeepSeek-AI, A.~Liu, and B.~F. et~al., ``Deepseek-v3 technical report,'' 2025. [Online]. Available: \url{https://arxiv.org/abs/2412.19437}
\BIBentrySTDinterwordspacing

\bibitem{zhang2025semantic}
R.~Zhang, H.~Qiao, P.~Xu, M.~Shang, and L.~Chen, ``Semantic-guided representation learning for multi-label recognition,'' \emph{arXiv preprint arXiv:2504.03801}, 2025.

\bibitem{hu2025co}
T.~Hu, W.~Zhang, J.~Guo, and H.~Li, ``Co-pseudo labeling and active selection for fundus single-positive multi-label learning,'' \emph{IEEE Transactions on Medical Imaging}, 2025.

\bibitem{wang2025splicemix}
L.~Wang, Y.~Zhan, L.~Ma, D.~Tao, L.~Ding, and C.~Gong, ``Splicemix: A cross-scale and semantic blending augmentation strategy for multi-label image classification,'' \emph{IEEE Transactions on Multimedia}, 2025.

\end{thebibliography}
}

\end{document}